\documentclass[10pt,twocolumn,letterpaper]{article}

\usepackage{cvpr}

\usepackage{longtable,booktabs,array,graphicx,tabularx,adjustbox}
\usepackage{calc}
\usepackage{microtype}
\usepackage{xurl}
\usepackage{placeins}
\usepackage{dblfloatfix}
\usepackage{flushend}

\DeclareUnicodeCharacter{2212}{\ensuremath{-}}
\DeclareUnicodeCharacter{2265}{\ensuremath{\geq}}
\DeclareUnicodeCharacter{2191}{\ensuremath{\uparrow}}
\DeclareUnicodeCharacter{2193}{\ensuremath{\downarrow}}

\makeatletter
\newcommand{\reportclearpage}{%
  \clearpage
  \global\@colht\textheight
  \global\@colroom\textheight
  \global\vsize\textheight
}
\makeatother

\definecolor{cvprblue}{rgb}{0.21,0.49,0.74}
\usepackage[breaklinks,colorlinks,allcolors=cvprblue]{hyperref}
\usepackage{graphicx}

\def\paperID{*****}
\def\confName{CVPR}
\def\confYear{2026}

\title{TRACE: Temporal Audit and Condition-aware Evaluation of Streaming Video Understanding}
\author{
Yibo Ma \quad Qianqian Zhang \quad Peng Liu \quad Tiancheng Zhao
\\[2pt]
\raisebox{-0.35\height}{\includegraphics[width=0.8cm]{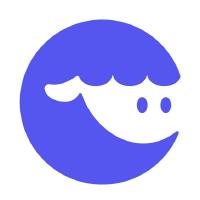}}
\hspace{0.1cm} Om AI Research
\\[2pt]
\texttt{tianchez@zju-bj.com}
}

\begin{document}
\maketitle
\begin{abstract}
Streaming video understanding requires models to interpret evidence as it arrives, yet current evaluations often report task scores without specifying when evidence becomes valid, how visual history is maintained, or how responses are triggered. As a result, similar scores may correspond to different workloads, failure modes, and operational behavior. We introduce TRACE (Temporal Audit and Condition-aware Evaluation), a condition-aware benchmark and evaluation framework that makes these factors explicit. TRACE combines temporally audited visual tasks with evidence timing and instruction-dependent trigger annotations, a unified causal Core--Adapter protocol that controls information availability while recording actual history processing and response events, and multidimensional reporting of answer quality, timeliness, response-selection behavior, workload, completion, and reliability. On 1,240 records from 517 videos, we evaluate eight publicly available models or systems in eight configurations. We find that nearly identical QA accuracy can mask substantial differences in completion, answer validity, and generation workload, while proactive performance separates into response quality, response delay, false alarms (responses emitted while no target window is currently valid and a later one remains), and missed target windows. These results show that streaming-video performance should be interpreted as execution-conditioned system behavior rather than a single score. Our benchmark and code can be accessed at \href{https://github.com/om-ai-lab/trace-bench}{https://github.com/om-ai-lab/trace-bench}.
\end{abstract}
\section{Introduction}
\label{sec:intro}

Multimodal large language models (MLLMs) for streaming video understanding receive information incrementally rather than as a complete video. They must process new observations as they arrive, retain information that may become useful later, and respond when a question is asked or when a monitored condition becomes true. This setting differs from conventional long-video understanding, where the complete video or a presampled context is typically available before inference. Consequently, models operating on a live stream cannot be compared meaningfully with offline video models without accounting for what evidence in video was available at the time of each response \citep{videollava2024}.

Causal access alone, however, does not fully specify a streaming evaluation. A reported score also depends on \emph{when} the supporting evidence becomes valid, \emph{how} the system maintains or reconstructs visual history, \emph{who} decides when a response is produced, and \emph{what} processing and failures occur along the way. Two systems can therefore obtain similar task scores while differing substantially in state maintenance, history replay, response timing, output redundancy, workload, completion, or answer validity. We argue that a streaming-video score is meaningful only when it is interpreted together with three forms of context: \textbf{temporal validity}, \textbf{execution conditions}, and \textbf{operational outcomes}.

We introduce TRACE (Temporal Audit and Condition-aware Evaluation), a benchmark and evaluation framework designed around this principle. TRACE combines temporally audited visual tasks with a unified causal Core--Adapter protocol. The task annotations specify when evidence supports a question-answering (QA) response and when a proactive response becomes valid; the execution protocol controls what video is available while recording how models process history and produce outputs; and the reporting layer measures task quality together with response timing, extra output, workload, completion, and runtime reliability under explicitly declared execution and deployment conditions.

We evaluate eight publicly available models or systems in eight configurations. The experiments show why the additional context matters: nearly identical QA accuracy can coincide with different completion rates, output volumes, and invalid-output rates, while similar proactive scores can mask large differences in response timing, false alarms, and missed target windows. Differences in history processing and evaluation boundary further show why a reported score must remain tied to the execution condition under which it was obtained.

Our contributions are threefold:
\begin{enumerate}[leftmargin=1.5em,label=(\arabic*),itemsep=0.25em,topsep=0.25em]
    \item \textbf{Condition-aware streaming evaluation.} We formulate streaming-video evaluation as a system-level measurement problem in which task scores are interpreted under explicit execution conditions. TRACE records visual-state maintenance, response initiation, and evaluation boundary instead of treating these choices as implicit properties of a model.
    \item \textbf{Temporally audited tasks and multidimensional measurement.} We construct a visual-only evaluation set from existing streaming-video benchmarks with reviewed evidence timing and instruction-dependent proactive trigger annotations. TRACE pairs these annotations with a unified Core--Adapter protocol that records actual execution behavior and reports quality, timeliness, extra output, workload, completion, and reliability.
    \item \textbf{Empirical analysis of current streaming systems.} Across eight public models or systems in eight configurations, we show that conventional task scores can conceal substantial operational differences, and that results obtained with different history-processing mechanisms or evaluation boundaries must be interpreted under their declared execution conditions rather than as directly interchangeable measurements.
\end{enumerate}

\section{Related Work}
\label{sec:related}

\paragraph{Long-video understanding.}
Long-video evaluations such as Video-MME \citep{videomme2025}, LongVideoBench \citep{longvideobench2024}, MLVU \citep{mlvu2024}, EgoSchema \citep{egoschema2023}, and LVBench \citep{lvbench2025} test event recognition, information aggregation, and temporal reasoning over long visual contexts. These benchmarks typically provide the complete video before answering, or construct the model context from the complete video in advance. They therefore measure long-context understanding without directly establishing whether a model can process evidence that arrives continuously or respond at the time an interaction requires it.

\paragraph{Streaming and online video evaluation.}
Recent benchmarks have established causal visibility as a central requirement for online video understanding. StreamingBench \citep{streamingbench2024} constrains access through question-arrival times, while RIVER \citep{river2026} and S-EMBER \citep{sember2026} further organize temporal relationships between questions, evidence, memory, and responses. OVBench \citep{videochatonline2025} evaluates online video understanding under causal input. These works establish that future information must be hidden, but causal access by itself does not specify how a system forms visual state or what processing occurs before a response is produced.

\paragraph{Proactive and interactive evaluation.}
A complementary line of work asks \emph{when} a model should respond. OVO-Bench \citep{ovobench2025} evaluates waiting for sufficient evidence, while ProactiveVideoQA \citep{proactivevideoqa2025}, OmniPro \citep{omnipro2026}, OmniMMI \citep{omnimmI2025}, and OmniInteract \citep{omniinteract2026} evaluate proactive or interactive response behavior under streaming input. These benchmarks motivate explicit response timing and event-dependent scoring. TRACE builds on this foundation but focuses on a narrower visual-only, single-instruction setting in order to jointly audit temporal validity, declare execution conditions, and measure operational behavior.

Table~\ref{tab:benchmark_comparison} summarizes the dimensions most relevant to TRACE before we introduce the benchmark in detail. The annotation columns distinguish instruction-dependent trigger rules, per-question evidence timing, and re-audited ground truth; the metric columns distinguish QA accuracy, response latency, answer parsability, Proactive accuracy, response timing, explicit false-alarm/miss diagnostics, workload, and completion. This comparison is intended to locate TRACE within the released evaluation landscape rather than to rank the underlying benchmarks.

\begin{table*}[!t]
\centering
\scriptsize
\setlength{\tabcolsep}{3.4pt}
\renewcommand{\arraystretch}{1.18}
\caption{Comparison with released streaming-video benchmarks most directly related to TRACE's scope. All listed projects provide public data and evaluation code with causal streaming access, verified in our 2026-07 release audit; a cross marks absence from the released definitions we audited, not a deficiency of the underlying work.}
\label{tab:benchmark_comparison}
\begin{tabular}{@{}l rr cc ccc ccc ccc cc@{}}
\toprule
& & & \multicolumn{2}{c}{\tiny Task} & \multicolumn{3}{c}{\tiny Annotation} & \multicolumn{3}{c}{\tiny QA metrics} & \multicolumn{3}{c}{\tiny Proactive metrics} & \multicolumn{2}{c}{\tiny Run-level} \\
\cmidrule(lr){4-5} \cmidrule(lr){6-8} \cmidrule(lr){9-11} \cmidrule(lr){12-14} \cmidrule(lr){15-16}
Benchmark & {\tiny Vid.} & {\tiny Rec.} & {\tiny QA} & \shortstack{\tiny Pro-\\ \tiny active} & \shortstack{\tiny Trigger\\ \tiny rules} & \shortstack{\tiny Evidence\\ \tiny timing} & \shortstack{\tiny Audited\\ \tiny ground truth} & \shortstack{\tiny Accu-\\ \tiny racy} & \shortstack{\tiny Latency\\ \tiny (QA)} & \shortstack{\tiny Invalid\\ \tiny outputs} & \shortstack{\tiny Accu-\\ \tiny racy} & \shortstack{\tiny Response\\ \tiny timing} & \shortstack{\tiny FA /\\ \tiny Miss} & \shortstack{\tiny Work-\\ \tiny load} & \shortstack{\tiny Comple-\\ \tiny tion} \\
\midrule
StreamingBench \citep{streamingbench2024} & 900 & 4,500 & \checkmark & \checkmark & $\times$ & $\times$ & $\times$ & \checkmark & $\times$ & $\times$ & \checkmark & $\times$ & $\times$ & $\times$ & $\times$ \\
OVO-Bench \citep{ovobench2025} & 644 & $\sim$3.1k & \checkmark & \checkmark & $\times$ & $\times$ & $\times$ & \checkmark & $\times$ & $\times$ & \checkmark & \checkmark & $\times$ & $\times$ & $\times$ \\
OVBench \citep{videochatonline2025} & 1,463 & 4,874 & \checkmark & $\times$ & $\times$ & $\times$ & $\times$ & \checkmark & $\times$ & $\times$ & $\times$ & $\times$ & $\times$ & $\times$ & $\times$ \\
RIVER \citep{river2026} & 1,067 & 4,278 & \checkmark & \checkmark & $\times$ & \checkmark & $\times$ & \checkmark & $\times$ & $\times$ & \checkmark & \checkmark & $\times$ & $\times$ & $\times$ \\
ProactiveVideoQA \citep{proactivevideoqa2025} & 1,377 & 1,427 & \checkmark & \checkmark & $\times$ & $\times$ & $\times$ & \checkmark & $\times$ & $\times$ & \checkmark & \checkmark & $\times$ & $\times$ & $\times$ \\
OmniPro \citep{omnipro2026} & 1,262 & 2,700 & \checkmark & \checkmark & $\times$ & $\times$ & $\times$ & \checkmark & $\times$ & $\times$ & \checkmark & $\times$ & $\times$ & $\times$ & $\times$ \\
OmniMMI \citep{omnimmI2025} & 1,121 & 2,290 & \checkmark & \checkmark & $\times$ & $\times$ & $\times$ & \checkmark & $\times$ & $\times$ & \checkmark & $\times$ & $\times$ & $\times$ & $\times$ \\
OmniInteract \citep{omniinteract2026} & 250 & 1,430 & \checkmark & \checkmark & $\times$ & $\times$ & $\times$ & \checkmark & $\times$ & $\times$ & \checkmark & \checkmark & \checkmark & $\times$ & $\times$ \\
\midrule
\textbf{TRACE (this paper)} & \textbf{517} & \textbf{1,240} & \checkmark & \checkmark & \checkmark & \checkmark & \checkmark & \checkmark & \checkmark & \checkmark & \checkmark & \checkmark & \checkmark & \checkmark & \checkmark \\
\bottomrule
\end{tabular}
\end{table*}

TRACE's In-window Accuracy is represented by the Proactive Accuracy column, while Median Response Delay is represented by Response timing. The FA/Miss column is reserved for releases that expose explicit response-selection error diagnostics (for example, false-positive/false-negative or false-alarm/miss behavior) rather than only folding those errors into an aggregate proactive score. Redundant output is not used as a cross-benchmark column because its semantics differ substantially across interaction protocols; TRACE reports it separately as a descriptive appendix statistic. Appendix~\ref{subsec:release-landscape} gives the wider release landscape and the criteria used in our audit.

\paragraph{Positioning of TRACE.}
Prior work increasingly enforces causal access and response timing, but these controls alone do not make streaming evaluations directly comparable. The same reported metric can still be computed under different evidence boundaries, history mechanisms, response-triggering protocols, and timing boundaries. TRACE makes these conditions explicit and evaluates them together with task quality and operational behavior.

\section{TRACE Evaluation Framework}
\label{sec:framework}

TRACE is designed around a simple premise: a streaming-video score is under-specified unless the evaluation also states \emph{when the evidence becomes valid}, \emph{under what execution conditions the response is produced}, and \emph{what operational behavior accompanies the score}. The framework therefore links temporal validity, execution conditions, and operational outcomes rather than reporting them as independent implementation details.

\subsection{Design Principles and Overview}

\paragraph{Temporal validity.}
The evaluation must establish which past visual evidence supports an answer and when a response becomes eligible. For QA, this requires evidence timing relative to question arrival. For proactive tasks, it requires instruction-dependent trigger semantics rather than a single generic event timestamp.

\paragraph{Explicit execution conditions.}
Models can satisfy the same causal-visibility rule while processing the stream differently. TRACE separates how visual history is maintained, whether proactive output is self-initiated, and what components are included in the evaluated system boundary. These categories define comparison conditions rather than capability levels.

\paragraph{Operational outcomes.}Task quality is reported together with timing, extra output, workload, completion, and runtime reliability. This allows a score to be interpreted together with the behavior and execution volume observed in the run, rather than as a self-contained scalar.

\paragraph{Framework components.}
TRACE operationalizes these principles through four connected components. First, a temporally audited visual task set supplies reviewed QA evidence times and proactive trigger annotations. Second, an Evaluation Core delivers timestamped frames and tasks on a controlled causal timeline. Third, model-specific Adapters connect the shared input protocol to heterogeneous models and systems while recording actual submissions, replay, state maintenance, outputs, and failures. Fourth, task-specific scoring converts those observations into quality, timeliness, behavior, workload, and reliability measurements under declared execution conditions.

\begin{figure*}[t]
\centering
\includegraphics[width=\textwidth,height=0.47\textheight,keepaspectratio]{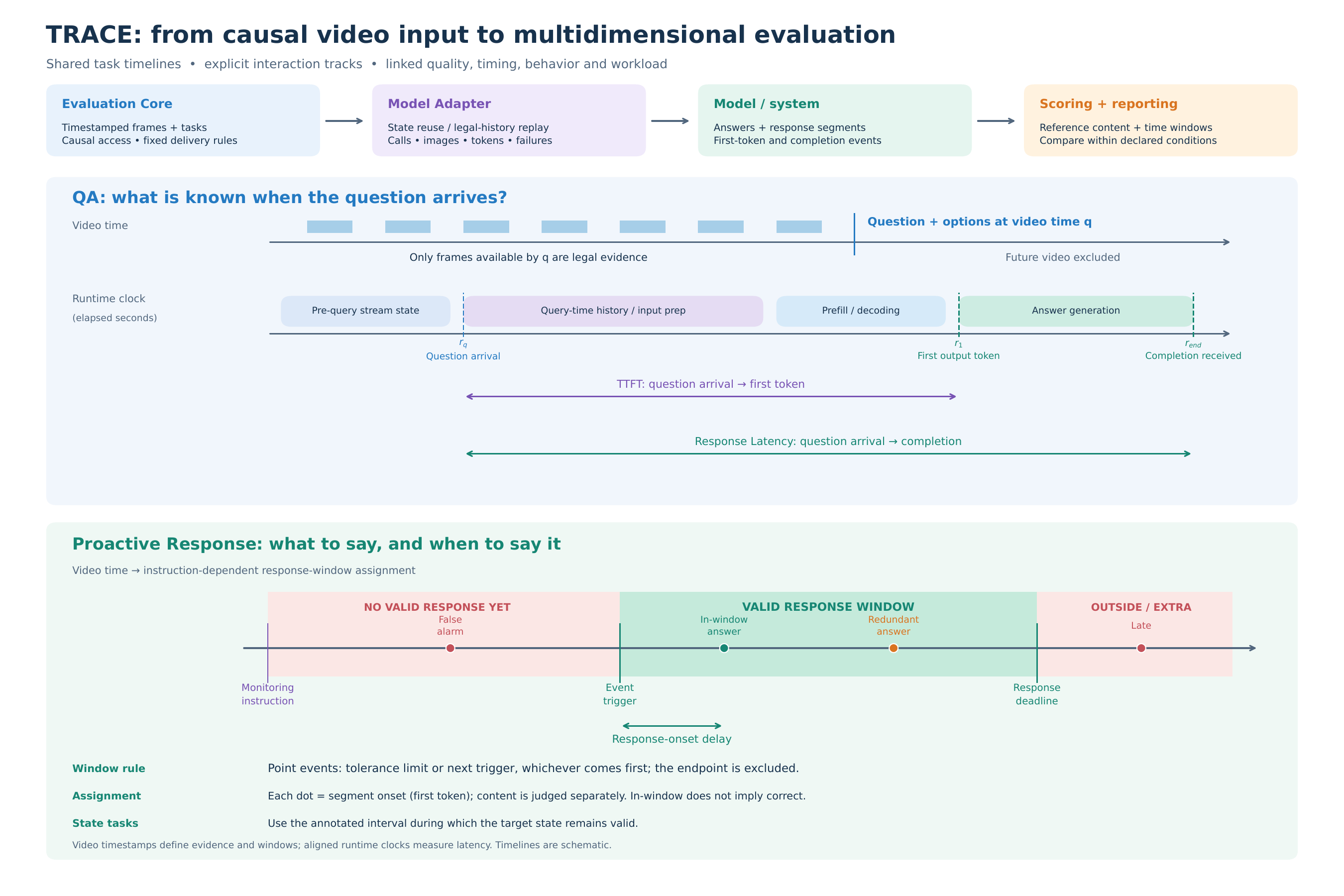}
\caption{TRACE overview from causal video input to condition-aware measurement. The top row connects the controlled Core, model-specific Adapter, evaluated model or system, and scoring layer. The middle timeline separates QA video-time evidence availability from runtime response events; the bottom timeline separates proactive trigger validity, response onset, and response content. These observation points support the quality, timing, extra-output, workload, completion, and reliability measurements reported by TRACE. Timelines are schematic.}
\label{fig:trace_overview}
\end{figure*}

Figure~\ref{fig:trace_overview} is the reading guide for the framework and for the metrics defined below. It links the execution path to the QA and Proactive Response timelines so that each reported measurement can be traced to an explicit observation point.

The middle timeline in Figure~\ref{fig:trace_overview} illustrates QA using separate video-time and runtime boundaries. The video-time question timestamp \(q\) determines which frames constitute legal evidence. On the runtime clock, \(r_q\) denotes \emph{question arrival}: the moment the question becomes active and the evaluated response path begins. The first output token occurs at \(r_1\), and completion is received at \(r_{\mathrm{end}}\). TRACE therefore uses the same runtime origin for both QA timing metrics: \emph{Time to First Token (TTFT)} is measured from \(r_q\) to \(r_1\), while \emph{Response Latency} is measured from \(r_q\) to \(r_{\mathrm{end}}\). Any query-time history reconstruction, input preparation, or queueing that occurs after question arrival is included in both intervals; Response Latency additionally includes generation after the first token.

The bottom timeline in Figure~\ref{fig:trace_overview} illustrates Proactive Response. An audited trigger or state interval defines when a response becomes eligible; the first-token onset assigns a generated segment to a response window, while response content is judged separately. \emph{In-window Accuracy} credits content only when it is assigned to a valid target window, the \emph{Median Response Delay} reports how quickly assigned responses begin, the \emph{False-alarm Rate} measures cases in which the system speaks when it should not yet speak, and the \emph{Miss Rate} measures target windows that receive no response. The figure therefore connects temporal annotation, runtime observation, and the four classes of reported outcomes: quality, timing, response behavior, and workload/reliability.

\begin{figure*}[t]
\centering
\includegraphics[width=\textwidth,height=0.43\textheight,keepaspectratio]{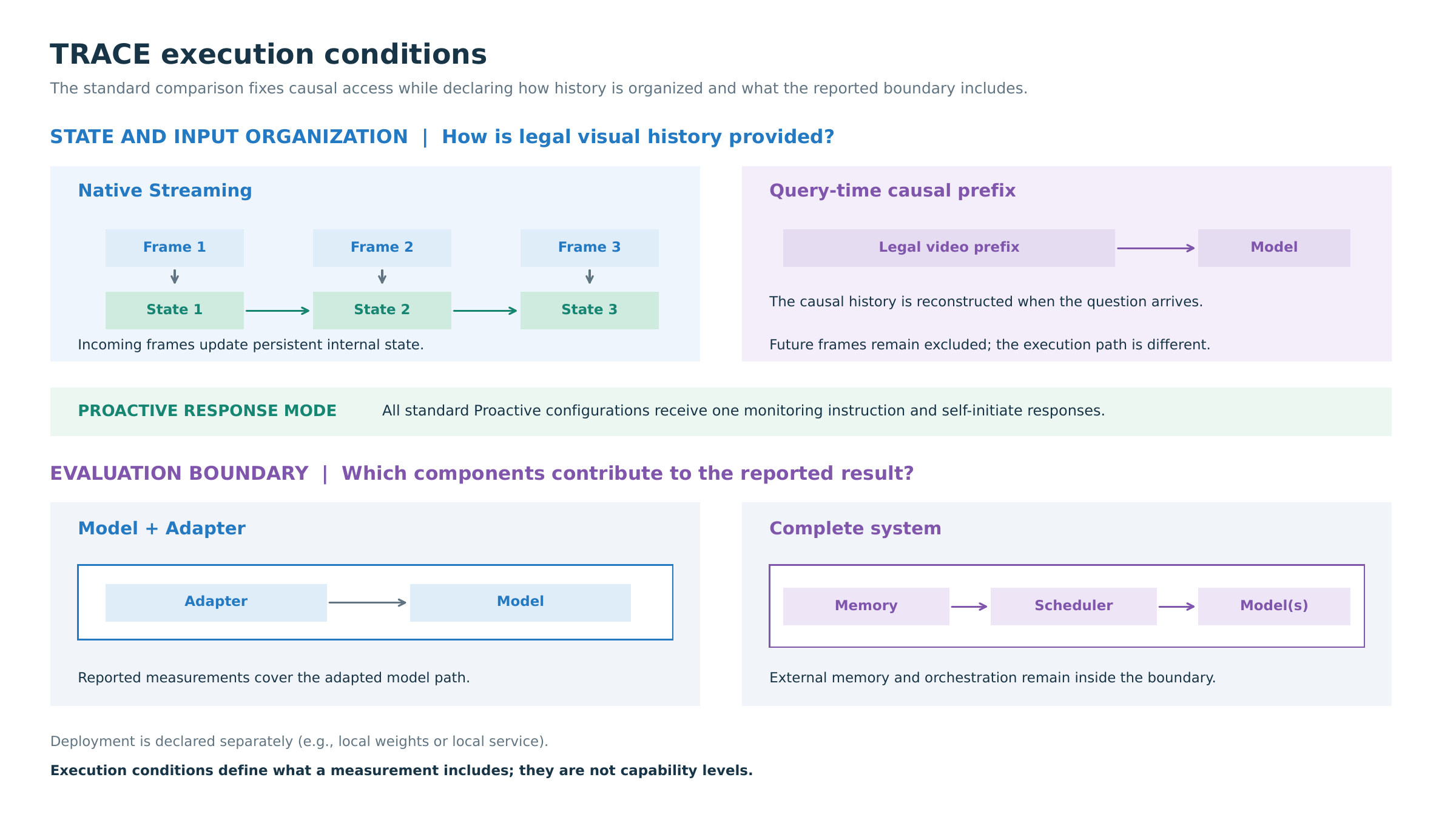}
\caption{Execution conditions recorded by TRACE. Native Streaming updates persistent state as frames arrive, whereas Non-native Streaming reconstructs a legal causal prefix or window at query time. Model + Adapter and complete-system boundaries specify which components contribute to reported measurements; deployment is declared separately. Proactive configurations in the standard comparison receive one monitoring instruction and self-initiate their responses. These properties define measurement conditions, not capability levels.}
\label{fig:execution_dimensions}
\end{figure*}

\subsection{Unified Causal Execution and Declared Conditions}

Core delivers timestamped video frames and tasks on a common timeline, using fixed sampling and image-processing rules. Models can access only video already received. Adapters preserve question, option, and instruction content while connecting this shared input to model-specific interfaces. They record actual image submissions, history replay, frame drops, state maintenance, outputs, and failures.

The protocol standardizes causal visibility and task arrival, but it does not force heterogeneous systems into the same internal state mechanism. Instead, TRACE records the relevant execution choices so that measured quality can be interpreted together with how the run was produced. In the top row of Figure~\ref{fig:trace_overview}, this distinction appears as a controlled Core feeding model-specific Adapters: the Core defines the legal stream, whereas the Adapter records what the evaluated configuration actually does with that stream.

For each run, TRACE records how visual history is maintained, how proactive output is initiated, and what components are included in the evaluation boundary. \textbf{Native Streaming} continuously receives frames and reuses persistent internal state, whereas \textbf{Non-native Streaming} reconstructs a legal causal prefix or window when a query arrives. In the standard Proactive comparison used in this paper, configurations receive one monitoring instruction and self-initiate their responses. The evaluation boundary distinguishes a model with its Adapter from a complete interaction system that may include external memory, scheduling, or multiple services. Deployment is declared separately, such as directly loaded weights or a local service.

Figure~\ref{fig:execution_dimensions} focuses on the two execution dimensions that vary in the reported experiments: \emph{state and input organization}, and the \emph{measurement boundary}. The former determines whether the legal visual history is maintained incrementally or reconstructed at query time; the latter determines whether reported measurements cover a model with its Adapter or an end-to-end system with additional memory, scheduling, or services. System-level timing and workload retain these tested boundaries and should not be read as hardware-neutral efficiency rankings.

\subsection{Evaluation Metrics}

The metric families in Table~\ref{tab:main_metrics} attach measurements to the same execution trace. QA metrics separate answer quality and format validity from runtime timing and workload. Proactive metrics separate content credit from response timeliness and from two distinct response-selection failures: speaking when no target response is yet valid, and failing to cover a target window. In particular, the main proactive quality metric is \emph{In-window Accuracy}, timeliness is summarized by the \emph{Median Response Delay}, and response behavior is summarized by the \emph{False-alarm} and \emph{Miss rates}. The arrows in the table indicate preferred directions only when the task population, completion conditions, and measurement boundaries are comparable.

\begin{table*}[t]
\centering
\footnotesize
\setlength{\tabcolsep}{4pt}
\renewcommand{\arraystretch}{1.13}
\caption{Main evaluation metrics. Arrows indicate preferred directions when task population, completion conditions, and measurement boundaries are comparable.}
\label{tab:main_metrics}
\begin{tabularx}{\textwidth}{@{}>{\raggedright\arraybackslash}p{0.17\textwidth} >{\raggedright\arraybackslash}p{0.31\textwidth} X@{}}
\toprule
\textbf{Task / dimension} & \textbf{Metric} & \textbf{Definition} \\
\midrule
QA quality & Accuracy $\uparrow$ & Correct, uniquely parsed records divided by all records \\
QA reliability & Completion Rate $\uparrow$ & Successfully completed records divided by all QA records \\
QA response & Response Latency $\downarrow$ & Question arrival \(r_q\) to completion receipt \(r_{\mathrm{end}}\) \\
QA response & Time to First Token (TTFT) $\downarrow$ & Question arrival \(r_q\) to first output token \(r_1\) \\
QA workload & Submitted Image Count $\downarrow$ & Actual image submissions, including repeats \\
QA workload & Total Output Tokens $\downarrow$ & All tokens generated during the QA run; Section~\ref{sec:experiments} reports the portion observed through interface telemetry \\
QA format & Invalid Output Rate $\downarrow$ & Fraction not uniquely mapped to a legal option \\
\midrule
Proactive quality & In-window Accuracy $\uparrow$ & Average score over all target windows, including partial credit \\
Proactive timeliness & Median Response Delay $\downarrow$ & Median trigger-to-onset delay over answered windows \\
Proactive behavior & False-alarm Rate $\downarrow$ & False-alarm response episodes divided by all assembled response episodes (global ratio) \\
Proactive behavior & Miss Rate $\downarrow$ & Target windows without an assigned response, divided by all target windows \\
Proactive reliability & Completion Rate $\uparrow$ & Successfully completed records divided by all Proactive records \\
\bottomrule
\end{tabularx}
\end{table*}

\paragraph{QA evaluation.}
A  question and its options arrive at video time \(q\). Only frames available by \(q\) are legal evidence; future video is excluded. As shown in the middle timeline of Figure~\ref{fig:trace_overview}, the runtime clock starts at question arrival \(r_q\). The first output token occurs at \(r_1\), and completion is received at \(r_{\mathrm{end}}\).

Accuracy counts a record as correct only when the parser recovers one unambiguous valid option:
\[
\mathrm{Accuracy}
=\frac{1}{N}\sum_{i=1}^{N}\mathbf{1}[\hat{y}_i=y_i].
\]
Failures, timeouts, and invalid outputs remain in the denominator. Completion Rate reports whether execution finishes successfully, while Invalid Output Rate reports responses for which no unique option can be recovered.

TRACE uses \textbf{Response Latency} as the primary QA timing metric:
\[
L_{\mathrm{comp}}=r_{\mathrm{end}}-r_q,
\]
where \(r_q\) denotes question arrival and \(r_{\mathrm{end}}\) denotes completion received by the Evaluation Core. It includes query-time history replay, input preparation, queueing, and generation after the question becomes active.

We additionally record \textbf{Time to First Token (TTFT)},
\[
\mathrm{TTFT}=r_1-r_q,
\]
as a diagnostic of when generation begins. TTFT is reported in Appendix~\ref{subsec:timing-workload} where available.

Submitted Image Count and Total Output Tokens describe the execution volume observed by the Model Adapter. They are workload measurements rather than normalized compute-cost measures. Parsing, timing coverage, and resource-accounting details are provided in Appendix~\ref{subsec:timing-workload}.

\paragraph{Proactive Response evaluation.}
Each Proactive Response task provides a monitoring instruction and one or more annotated event triggers or valid state intervals. As shown in the bottom timeline of Figure~\ref{fig:trace_overview}, each trigger defines a valid response window. Before that window, there is no valid response yet; responses beginning during the window can be assigned to the target, while responses outside it do not receive strict-window credit.

For a point trigger \(t_i\), tolerance \(W\), and next trigger \(t_{i+1}\), TRACE defines
\[
[t_i,\min(t_i+W,t_{i+1})).
\]
The endpoint is excluded. State tasks instead use their annotated valid intervals. Main results use \(W=5\) seconds.

Responses are assigned by segment onset, i.e., the first output token. Timing determines which response window a segment belongs to, while response content is scored separately. An in-window answer is therefore not necessarily correct.

Sequential Steps Recognition (SSR) and Clues Reveal Responding (CRR) use a semantic Judge; the remaining tasks use normalized exact matching. In-window Accuracy is the sum of assigned window scores divided by all target windows.

Timeliness is measured by \textbf{Median Response Delay}, the median response-onset delay from the event trigger to the first output token of the assigned response.

TRACE also reports two response-selection errors. A \textbf{false alarm} is a response that begins when no response window is currently valid and a later response opportunity remains. The global False-alarm Rate is
\[
\mathrm{FA}
=
\frac{\sum \text{false-alarm response episodes}}
{\sum \text{all assembled response episodes}}.
\]
The \textbf{Miss Rate} is the fraction of target windows that receive no assigned response.

Thus, In-window Accuracy measures response quality, Median Response Delay measures response timing, and FA and Miss measure response-selection behavior. Appendix~\ref{subsec:timing-workload} reports timing coverage and additional output statistics.

\subsection{Evaluation Scope}
TRACE currently evaluates visual-only, single-instruction streaming video understanding. QA tasks introduce a question at a specified video time, while Proactive Response tasks provide a monitoring instruction before one or more target events or states. The standard protocol uses causal visual access at 1 FPS. Proactive configurations in the standard comparison self-initiate responses after one monitoring instruction, and execution boundaries are reported explicitly. Section~\ref{sec:benchmark} describes task construction and temporal annotation; Section~\ref{sec:experiments} reports the evaluated models and results.

\section{Benchmark Construction and Temporal Audit}
\label{sec:benchmark}

TRACE builds its evaluation set from visual-only tasks in StreamingBench
\citep{streamingbench2024} and OVO-Bench \citep{ovobench2025}. 
We retain tasks that can be answered from visual evidence without audio, subtitles, or external knowledge. Questions, instructions, answers, and timestamps are converted into a common data structure while preserving their source identities. We organize the selected tasks into QA and Proactive Response, audit their temporal annotations, and form the standard set used for 1 FPS evaluation.

\subsection{Temporal Annotations for Causal Evaluation}

QA annotations record the question-arrival time and the visual evidence supporting the answer. During review, annotators watched each video, located the relevant evidence, and marked the earliest time at which that evidence supported the answer given the question and options, in the spirit of temporal moment localization \citep{tan2d2020, qvhighlights2021}. This time cannot be later than question arrival. The interval between the two times is the evidence-to-question distance. The distance is not, by itself, a difficulty measure: repeated evidence and content complexity can also affect the task. For questions labeled unanswerable, the permitted history before question arrival must be checked.

Proactive annotations follow the condition specified by the instruction. For event-onset tasks, the trigger is the first time the target condition holds. For event-completion tasks, it is the end of the requested action. For sufficient-evidence tasks, it is the earliest time at which the visible clues support a unique answer and rule out the relevant alternatives. State tasks use the interval during which the answer remains valid, ending when the state ceases to hold or is replaced.

These four types describe instruction-dependent trigger rules, not mutually exclusive types of video content. The same action may use an onset or completion trigger depending on the instruction. Each annotation also records the expected answer and event order, with the instruction preceding the target trigger. The annotations define video-grounded reference times; the response tolerance used for scoring is specified in Section~\ref{sec:framework}. Examples are given in Appendix~\ref{subsec:trigger-examples}.

\subsection{Data Preparation and Quality Control}

The construction workflow combined manual organization, model-assisted risk screening, and human review. Model screening was used to identify possible problems with visual answerability, answer uniqueness, evidence timing, and proactive trigger boundaries. Annotators then inspected the source videos, reviewed the questions or instructions, answers, and temporal annotations, and either corrected the records or removed records that could not be repaired reliably. High-risk records received focused review, while lower-risk records were sampled according to the review procedure.

Researchers performed targeted verification on records that remained unresolved after the annotators' second pass or were later flagged against the evaluation criteria. They did not independently re-annotate every record. Annotators could see the existing annotations and model opinions, so this was assisted quality control rather than blind duplicate annotation. The release publishes aggregate review statistics, a per-record change ledger with prior and revised values, and an exclusion list with reasons. In release set, the ledger contains 328 edited records comprising 952 field-level changes, together with 52 excluded records. Of the 328 edited records, 325 change at least one timing or trigger field: 100/103 edited QA records and all 225 edited Proactive records. For directly comparable numeric timestamp fields, the median absolute revision is 3.0~s for QA (127 field changes; 90th percentile 58.2~s) and 2.24~s for Proactive Response (150 field changes; 90th percentile 20.76~s). These statistics quantify annotation movement rather than model-score changes, but they show that the temporal audit is not cosmetic: a revised QA evidence time changes the legal causal history associated with a question, while a revised proactive trigger changes the boundary used for in-window credit, false alarms, and misses. The internal review workspace, including model-audit runs and annotator session history. Section~\ref{subsec:standard-set} separates records with a retained human decision in this review round from records retained after model screening alone.

\subsection{Final Standard Set and Sampling Observability}
\label{subsec:standard-set}

The released dataset contains 1,248 records over 522 videos: 833 QA records and 415 Proactive Response records with 1,338 response windows. One video is shared across the two task types after deduplication by source and path.

The standard 1 FPS evaluation set contains 1,240 records over 517 videos: all 833 QA records and 407 Proactive Response records with 1,270 response windows. Eight Proactive records containing 25 windows of at most one second are retained as a released sampling-stress subset but excluded from the primary evaluation because they cannot be reliably sampled at 1 FPS.

Before release, we checked record IDs, answer-option consistency, temporal ordering, response-window boundaries, media bounds, and source mappings. The released annotations, subset definitions, and change ledger are sufficient to reproduce the evaluation population and its revision provenance; the internal review workspace is not required for execution.

\section{Experiments}
\label{sec:experiments}

We evaluate eight publicly available models or systems in eight configurations on the TRACE standard set.  Failed runs remain in the corresponding denominators. This section reports the experimental setup and the main measured results; Section~\ref{sec:analysis} analyzes what these results imply for streaming-system comparison.

\begin{table*}[tb]
\centering
\small
\setlength{\tabcolsep}{4pt}
\renewcommand{\arraystretch}{1.15}
\caption{Per-model configurations and result boundaries.}
\label{tab:configurations}
\begin{tabular}{@{}
  >{\raggedright\arraybackslash}p{(\linewidth - 6\tabcolsep) * \real{0.2500}}
  >{\raggedright\arraybackslash}p{(\linewidth - 6\tabcolsep) * \real{0.2500}}
  >{\raggedright\arraybackslash}p{(\linewidth - 6\tabcolsep) * \real{0.2500}}
  >{\raggedright\arraybackslash}p{(\linewidth - 6\tabcolsep) * \real{0.2500}}@{}}
\toprule\noalign{}
\begin{minipage}[b]{\linewidth}\raggedright
Model/system
\end{minipage} & \begin{minipage}[b]{\linewidth}\raggedright
State and input organization
\end{minipage} & \begin{minipage}[b]{\linewidth}\raggedright
Proactive trigger
\end{minipage} & \begin{minipage}[b]{\linewidth}\raggedright
Deployment and evaluation boundary
\end{minipage} \\
\midrule\noalign{}

AURA & QA: Non-native prefix processing at query time; Proactive: persistent history with incremental visual-state updates & autonomous & Local service; model + Adapter \\
MOSS-VL & Native; real-time session and asynchronous frame queue & autonomous & Local weights; model + Adapter \\
MOSS-Preview & Native; real-time session and asynchronous frame queue & autonomous & Local weights; model + Adapter \\
LiveCC & Native; persistent session and streaming generation & autonomous & Local weights; model + Adapter \\
ThinkStream & Native; two-frame blocks and persistent state & autonomous & Local weights; model + Adapter \\
VideoLLM-Online & Native; per-frame visual representations and persistent key-value (KV) state & autonomous & Local weights; model + Adapter \\
MiniCPM-O (native duplex) & Native; per-frame input and persistent duplex state & autonomous & Local weights; model + Adapter \\
JoyAI & Current segments and external summary memory; model-level incremental state unverified & autonomous & Local multiple services; end-to-end system \\

\bottomrule
\end{tabular}
\end{table*}

\subsection{Experimental Setup and Comparison Groups}

Core delivers timestamped red-green-blue (RGB) frames at 1 FPS with real-time pacing and controls question or instruction arrival. QA uses the same questions and answer parser across configurations; semantic Proactive scoring uses the same Qwen3.5-35B-A3B Judge configuration (Appendix~\ref{subsec:judge-calibration}). The reported local-model runs were executed on NVIDIA A100 80GB GPUs; exact per-run checkpoints, Adapter settings, and resolved runtime configurations are released with the project. Timing and workload measurements remain configuration-specific rather than hardware-normalized.

Table~\ref{tab:configurations} makes the execution conditions of every tested configuration explicit before any scores are compared. It separates visual state maintenance, proactive response mode, and deployment/evaluation boundary. The model + Adapter configurations include AURA \citep{aura2026}, LiveCC \citep{livecc2025}, MOSS-Preview \citep{mossvideopreview2026}, MOSS-VL \citep{mossvl2026}, ThinkStream \citep{thinkstream2026}, VideoLLM-Online \citep{videollmonline2024}, and MiniCPM-O 4.5 \citep{minicpmo45_2026}. AURA reconstructs the QA prefix at query time and maintains persistent state for Proactive Response; the other model-side configurations use their tested native streaming interfaces. JoyAI \citep{joyai2026} is evaluated as a complete system. All Proactive configurations in the main comparison receive one monitoring instruction and self-initiate their responses.

The tables report the tested configurations rather than hardware-normalized performance. Completion is reported at three granularities with separate denominators---QA record level, Proactive record level, and stream level (Appendix~\ref{subsec:timing-workload})---and the three are not directly comparable.

\subsection{Main QA Results}

\begin{figure*}[!tb]
\centering
\includegraphics[width=\textwidth,height=0.65\textheight,keepaspectratio]{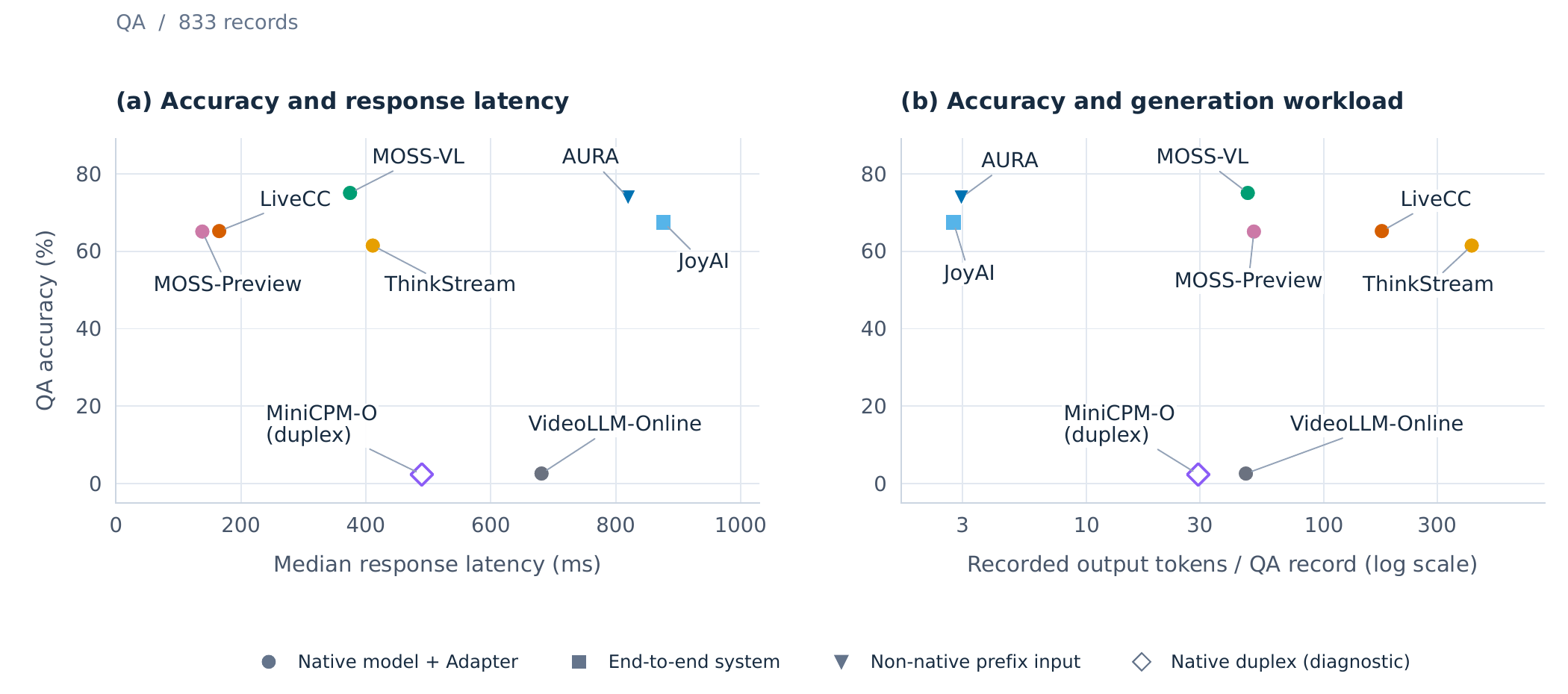}
\caption{QA accuracy alongside median response latency (left) and recorded output tokens per scheduled QA record (right, logarithmic axis).  Shapes identify native model + Adapter, complete-system, non-native prefix-input, and native duplex configurations. In both panels, higher accuracy together with lower latency or lower generation volume appears toward the upper left. The axes report observed measurements under the boundaries in Table~\ref{tab:configurations}, not hardware-normalized efficiency.}
\label{fig:qa_accuracy_workload}

\vspace{0.4em}
\small
\setlength{\tabcolsep}{4pt}
\renewcommand{\arraystretch}{1.12}
\centering
\captionof{table}{QA results by execution condition. Latency is the median Response Latency in milliseconds; output tokens are observed run totals over the scheduled population and are not normalized by completion.}
\label{tab:qa_results}
\begin{tabular}{@{}
  >{\raggedright\arraybackslash}p{(\linewidth - 12\tabcolsep) * \real{0.1111}}
  >{\raggedleft\arraybackslash}p{(\linewidth - 12\tabcolsep) * \real{0.1481}}
  >{\raggedleft\arraybackslash}p{(\linewidth - 12\tabcolsep) * \real{0.1481}}
  >{\raggedleft\arraybackslash}p{(\linewidth - 12\tabcolsep) * \real{0.1481}}
  >{\raggedleft\arraybackslash}p{(\linewidth - 12\tabcolsep) * \real{0.1481}}
  >{\raggedleft\arraybackslash}p{(\linewidth - 12\tabcolsep) * \real{0.1481}}
  >{\raggedleft\arraybackslash}p{(\linewidth - 12\tabcolsep) * \real{0.1481}}@{}}
\toprule
\multicolumn{7}{l}{\textit{(a) Native models with Adapters}} \\
\midrule
\begin{minipage}[b]{\linewidth}\raggedright Model\end{minipage} &
\begin{minipage}[b]{\linewidth}\raggedleft Accuracy\end{minipage} &
\begin{minipage}[b]{\linewidth}\raggedleft Completion\end{minipage} &
\begin{minipage}[b]{\linewidth}\raggedleft Response Latency (median, ms)\end{minipage} &
\begin{minipage}[b]{\linewidth}\raggedleft Submitted images\end{minipage} &
\begin{minipage}[b]{\linewidth}\raggedleft Recorded output tokens\end{minipage} &
\begin{minipage}[b]{\linewidth}\raggedleft Invalid output\end{minipage} \\
\midrule
LiveCC & 65.19\% & 93.88\% & 165.2 & 21,917 & 146,064 & 6.12\% \\
MOSS-Preview & 65.07\% & 100.00\% & 138.1 & 30,835 & 42,206 & 1.92\% \\
MOSS-VL & 75.03\% & 99.88\% & 374.6 & 30,835 & 39,768 & 0.84\% \\
ThinkStream & 61.46\% & 100.00\% & 411.0 & 30,835 & 349,476 & 0.00\% \\
VideoLLM-Online & 2.64\% & 100.00\% & 681.1 & 30,835 & 39,047 & 95.32\% \\
MiniCPM-O (native duplex) & 2.40\% & 100.00\% & 489.5 & 30,835 & 24,591 & 93.52\% \\
\addlinespace
\midrule
\multicolumn{7}{l}{\textit{(b) End-to-end system}} \\
\midrule
JoyAI & 67.47\% & 91.36\% & 876.1 & 17,235 & 2,286 & 8.52\% \\
\addlinespace
\midrule
\multicolumn{7}{l}{\textit{(c) Non-native prefix-input}} \\
\midrule
AURA & 73.83\% & 99.88\% & 819.7 & 31,170 & 2,469 & 0.60\% \\
\bottomrule
\end{tabular}
\end{figure*}

Table~\ref{tab:qa_results} and Figure~\ref{fig:qa_accuracy_workload} report Accuracy together with completion, median Response Latency, image submissions, recorded output tokens, and invalid-output rate. Each QA record has one scheduled question. Figure~\ref{fig:qa_accuracy_workload} should be read as a set of paired operational views rather than as an efficiency leaderboard: the left panel places accuracy against the median response latency, while the right panel places the same accuracy values against observed output-token volume on a logarithmic axis. Marker shapes identify execution boundaries so that a point's position is not detached from how that result was obtained. 

Table~\ref{tab:qa_results} intentionally keeps these operational measurements separate rather than aggregating them into a single score. The paired view in Figure~\ref{fig:qa_accuracy_workload} already shows why: LiveCC and MOSS-Preview occupy nearly the same accuracy level, yet the table reveals different completion, output-token, and invalid-output profiles; MOSS-VL combines higher QA accuracy with a recorded token total similar to MOSS-Preview; and VideoLLM-Online's low accuracy coincides with a high invalid-output rate rather than incomplete execution. MiniCPM-O native duplex exposes a different interface failure mode, with 93.52\% of QA outputs not recoverable as a unique option under the text protocol. Section~\ref{sec:analysis} analyzes these contrasts as evidence for condition-aware reporting rather than treating any single axis as a model ranking.

\subsection{Main Proactive Results}
\label{subsec:proactive-results}

\begin{figure*}[t]
\centering

\includegraphics[
    width=0.75\textwidth,
    trim=10 5 10 5,
    clip
]{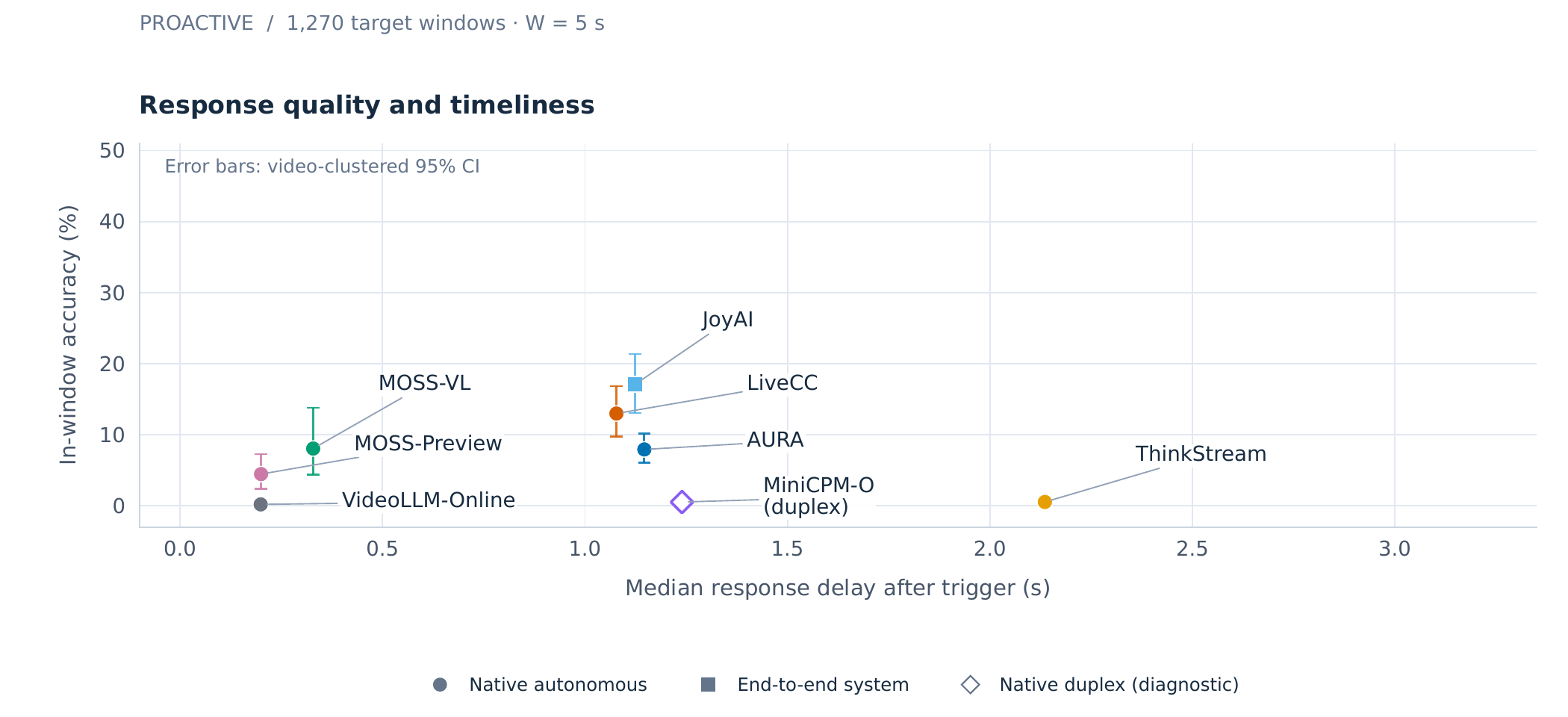}

\vspace{2pt}
\small (a) Quality and response timing

\vspace{6pt}

\includegraphics[
    width=0.75\textwidth,
    trim=10 5 10 5,
    clip
]{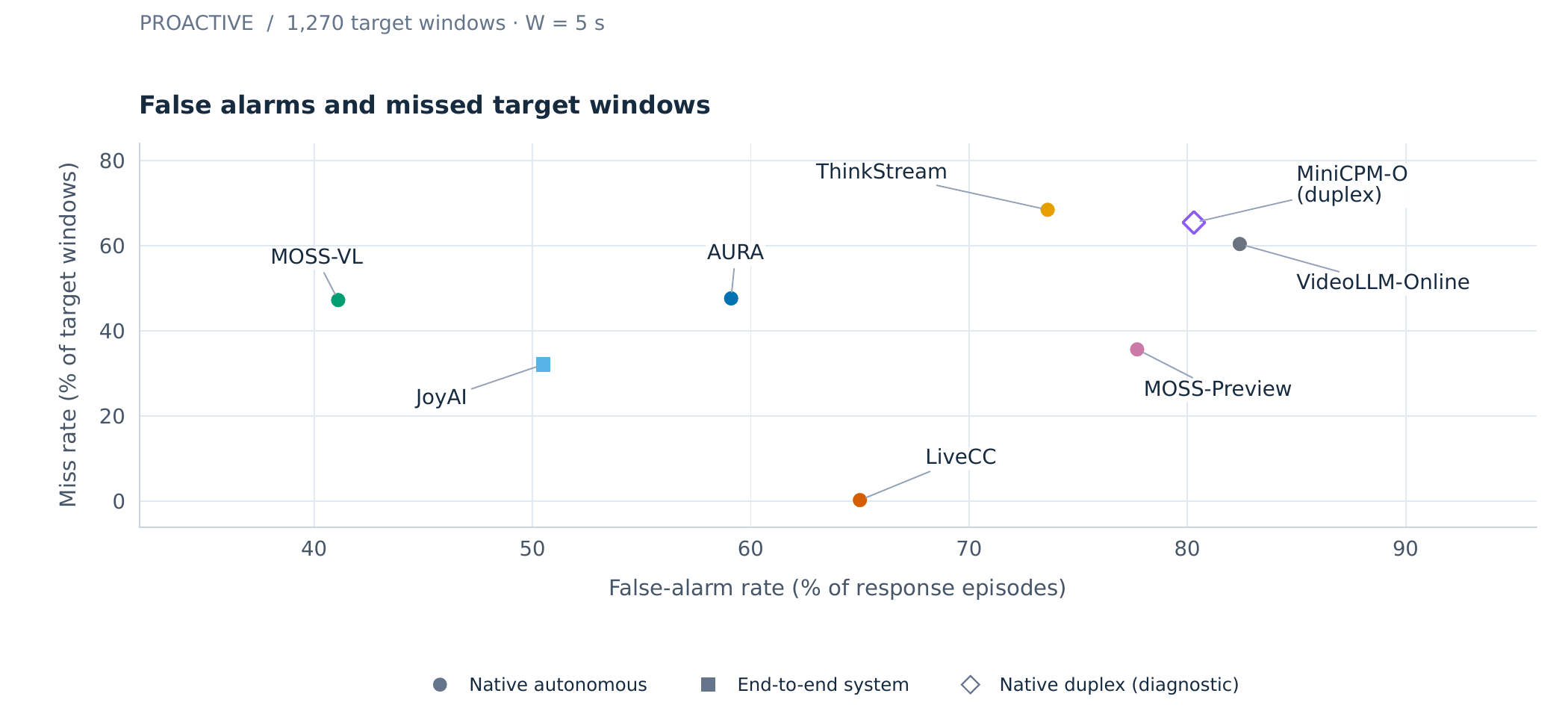}

\vspace{0.6pt}
\small (b) Response-selection behavior

\caption{Proactive Response results on the standard set.
(a) In-window Accuracy with video-clustered 95\% confidence intervals versus Median Response Delay. Higher accuracy and lower delay appear toward the upper left.
(b) Response-selection behavior, summarizing two complementary forms of response-selection error; lower values on both axes are preferred.
Marker shapes distinguish native model + Adapter, complete-system, and native duplex diagnostic configurations.}
\label{fig:proactive_results}
\end{figure*}

Table~\ref{tab:proactive_results} presents the self-initiated model + Adapter results and the complete-system result under their declared execution boundaries. In-window Accuracy averages content credit over target windows, the Median Response Delay reports how quickly observed assigned responses begin, the False-alarm Rate measures response episodes emitted when the system should not yet speak, and the Miss Rate measures target-window non-coverage. Figure~\ref{fig:proactive_results} (a) places quality against delay with video-clustered confidence intervals, and Figure~\ref{fig:proactive_results} (b) exposes the behavior contrast that a single quality score would hide.
\begin{figure*}[!tb]
\centering
\small
\setlength{\tabcolsep}{6pt}
\renewcommand{\arraystretch}{1.15}
\captionof{table}{Proactive results by response track. In-window Accuracy and Miss are window-level; False-alarm Rate is a global response-episode ratio; Median Response Delay is conditional on answered windows with an observed onset. Delay coverage is reported in Appendix~\ref{subsec:timing-workload}. Image and token columns are observed run totals over the scheduled population and follow the accounting boundary of Table~\ref{tab:qa_results}.}
\label{tab:proactive_results}
\begin{tabular}{@{}l rr rr rr@{}}
\toprule
& & & \multicolumn{2}{c}{\tiny Response behavior} & \multicolumn{2}{c}{\tiny Workload} \\
\cmidrule(lr){4-5} \cmidrule(lr){6-7}
\textbf{Model} & \textbf{In-window Acc} & \shortstack{\tiny Median\\ \tiny delay (s)} & \shortstack{\tiny False-\\ \tiny alarm} & \shortstack{\tiny Miss\\ \tiny rate} & \shortstack{\tiny Submitted\\ \tiny images} & \shortstack{\tiny Output\\ \tiny tokens} \\
\midrule
\multicolumn{7}{l}{\textit{(a) Autonomous model + Adapter}} \\
LiveCC & 12.98\% & 1.08 & 65.0\% & 0.31\% & 23,243 & 78,478 \\
AURA & 7.92\% & 1.15 & 59.1\% & 47.64\% & 439,400 & 50,957 \\
MOSS-VL & 8.05\% & 0.33 & 41.1\% & 47.24\% & 34,520 & 74,672 \\
MOSS-Preview & 4.45\% & 0.20 & 77.7\% & 35.67\% & 34,520 & 193,517 \\
ThinkStream & 0.52\% & 2.14 & 73.6\% & 68.43\% & 32,783 & 373,897 \\
VideoLLM-Online & 0.18\% & 0.20 & 82.4\% & 60.39\% & 32,783 & 42,162 \\
MiniCPM-O (native duplex) & 0.51\% & 1.24 & 80.3\% & 65.43\% & 32,617 & 23,636 \\
\addlinespace
\midrule
\multicolumn{7}{l}{\textit{(b) End-to-end system}} \\
JoyAI & 17.08\% & 1.12 & 50.5\% & 32.13\% & 313,658 & 797,954 \\
\bottomrule
\end{tabular}
\end{figure*}



Table~\ref{tab:proactive_results} likewise keeps proactive quality, timing, response-selection behavior, and workload separate rather than collapsing them into a single score. The paired view in Figure~\ref{fig:proactive_results} already shows why: MOSS-VL and AURA have nearly the same In-window Accuracy but differ substantially in False-alarm Rate and Median Response Delay, while LiveCC combines higher window-level accuracy and near-complete target-window coverage with a high False-alarm Rate. Median Response Delay is conditional on answered windows with an observed onset, so it describes observed response timing rather than target-window coverage. Section~\ref{sec:analysis} analyzes these contrasts as evidence for condition-aware reporting rather than treating any single axis as a model ranking.

\section{Analysis and Key Findings}
\label{sec:analysis}


The results reveal three reasons why a single task score is insufficient
for streaming-video evaluation. First, temporal audit materially changes
the boundaries on which correctness and response timing are measured.
Second, proactive performance separates into quality, timing, and
response-selection behavior rather than varying along one common axis.
Third, the same model or task score can have different meanings under
different execution conditions and system boundaries. We examine these
three findings below.


\subsection{Finding 1: Temporal Audit Establishes Reliable Measurement Boundaries}

TRACE's temporal audit is not merely a data-cleaning step: it determines
the boundaries on which both QA and Proactive Response are evaluated.
Across the reviewed data, 328 records were edited and 52 were excluded,
with 325 of the 328 edited records involving temporal or trigger-related
fields. For QA, 100 of 103 edited records contained timing revisions; for
Proactive Response, all 225 edited records revised or added trigger
annotations. Among directly comparable timestamps, the median absolute
revision was 3.0\,s for QA and 2.24\,s for Proactive Response, with much
larger shifts in the upper tail.

These revisions affect different parts of the two task types. In QA,
reviewed evidence times determine which visual information is legally
available before question arrival and how much history the answer requires.
In Proactive Response, audited triggers and valid intervals determine
whether a response is in-window, how response delay is measured, and
whether an unassigned response contributes to a false alarm or a target
window to a miss. The same model output can therefore receive different
evaluation outcomes under different temporal boundaries.

The audit thus provides the shared temporal basis on which TRACE's
downstream measurements are defined. Rather than treating timestamps as
fixed metadata, TRACE makes temporal validity an explicit, reviewed part
of the evaluation protocol.

\subsection{Finding 2: Similar Task Scores Can Hide Different System Behaviors}

Across both QA and Proactive Response, task quality does not determine
the operational or response behavior observed during evaluation.
TRACE therefore keeps these measurements separate instead of collapsing
them into a single aggregate score.

The QA results provide a simple example. LiveCC and MOSS-Preview achieve
nearly identical Accuracy, at 65.19\% and 65.07\%, yet their operational
profiles differ substantially. MOSS-Preview completes all records,
compared with 93.88\% completion for LiveCC, records 42,206 output tokens
rather than 146,064, and has a lower Invalid Output Rate
(1.92\% versus 6.12\%). Similar accuracy therefore does not imply similar
completion, answer validity, or generation workload. 

The same separation is more pronounced for Proactive Response, where
quality, timing, and response selection can vary independently. MOSS-VL
and AURA occupy essentially the same accuracy, with 8.05\% and
7.92\% In-window Accuracy. Their
response behavior, however, is not equally similar. MOSS-VL has a
41.1\% False-alarm Rate and a 0.33\,s median Response Delay, compared
with 59.1\% and 1.15\,s for AURA. Thus, two models can receive nearly the
same window-level quality score while differing substantially in when
and how often they respond outside valid opportunities.

The two MOSS models illustrate another form of separation.
MOSS-VL reaches 8.05\% In-window Accuracy with 74,672 recorded output
tokens, compared with 4.45\% and 193,517 tokens for MOSS-Preview.
MOSS-Preview begins its observed assigned responses slightly earlier at
the median (0.20\,s versus 0.33\,s), but it also produces a much higher
False-alarm Rate (77.7\% versus 41.1\%) while missing fewer target windows
(35.67\% versus 47.24\%). Faster observed responses and lower miss rates
therefore do not necessarily imply better response selection or higher
response quality. The measurements describe distinct aspects of behavior.

LiveCC makes this distinction especially clear. Its autonomous
In-window Accuracy is 12.98\%, and on the shared-video population it
exceeds AURA by $+$4.92 points (95\% CI $+$2.28 to $+$7.97). At the same
time, it misses only 0.31\% of target windows, but 65.0\% of its assembled
response episodes are false alarms. This is not a contradiction: Miss Rate
measures whether target windows receive an assigned response, whereas
False-alarm Rate measures responses emitted when no response window is
currently valid and a later opportunity remains. A configuration can
therefore cover nearly every target window while also producing many
responses at inappropriate times. Conversely, ThinkStream and
VideoLLM-Online combine high Miss Rates (68.43\% and 60.39\%) with high
False-alarm Rates (73.6\% and 82.4\%), showing that frequent off-window
output does not guarantee target-window coverage.

These cases show that accuracy,
completion, answer validity, workload, response timing, false alarms,
and misses expose different aspects of system behavior. TRACE reports
these quantities separately because no single one reliably predicts the
others. The resulting multidimensional view does not replace task
accuracy; it explains how a reported task score was obtained and which
operational or response behaviors accompany it.

\subsection{Finding 3: Execution Conditions Define What a Score Measures}

A reported metric remains tied to the execution path that produced it. This matters even when all Proactive configurations in the main comparison are self-initiated. AURA, for example, reconstructs the legal visual prefix at QA query time but maintains persistent incremental state for Proactive Response. Its QA latency and submitted-image count therefore describe a different history-processing path from native stateful configurations, rather than a hardware-normalized notion of model speed or efficiency.

Evaluation boundary creates a second distinction. JoyAI's 17.08\% In-window Accuracy is measured at the complete-system boundary, including the tested memory and scheduling components, whereas the model + Adapter results measure the model-side execution path exposed through TRACE. These measurements remain informative, but the same numeric metric does not imply that the same components, workload, or failure sources are included in the result.

MiniCPM-O native duplex provides a complementary interface diagnostic. It is evaluated through its native persistent listen/speak interaction, but its spoken-style QA outputs are poorly matched to the benchmark's text-option protocol: 93.52\% cannot be recovered as a unique option, yielding 2.40\% QA Accuracy. On Proactive Response it remains self-initiated, with 0.51\% In-window Accuracy and a 65.43\% Miss Rate. These numbers are reported as measurements of the tested interface, not as interface-independent estimates of the underlying model's capability. Together, the examples show why TRACE keeps history organization, response mode, and system boundary visible when interpreting a score.

Taken together, these findings show that temporal validity, multidimensional behavior, and execution conditions each affect how a streaming result should be interpreted. TRACE makes these distinctions explicit so that similar task scores are not mistaken for equivalent system behavior.

\section{Conclusion}
\label{sec:conclusion}



TRACE reframes streaming video understanding as a condition-aware measurement problem rather than a single-score task. A streaming result is interpretable only when three elements are made explicit together: the temporal validity of the supporting evidence, the execution conditions under which responses are produced, and the operational behavior accompanying task quality. TRACE operationalizes this view through temporally audited QA evidence and proactive response windows, a controlled Core–Adapter protocol, and multidimensional reporting of quality, timing, response-selection behavior, workload, completion, and reliability.

The experiments show that these distinctions are consequential. Temporal audit materially revises the boundaries on which QA evidence and proactive responses are evaluated; similar task scores can accompany substantially different completion, answer validity, workload, response timing, false alarms, and missed target windows; and differences in history organization or complete-system boundaries change what a reported score represents. TRACE therefore improves interpretability not by replacing task accuracy, but by making the temporal, operational, and execution conditions behind that accuracy explicit and reproducible.

\section{Limitations and Future Directions}
\label{sec:limitations}

\paragraph{Limitations.}
The current findings are limited to the tested models or systems on visual-only, single-instruction tasks at 1 FPS. Proactive tasks emphasize positive triggers, so False-alarm Rate measures responses emitted when no response window is currently valid and a later response opportunity remains, rather than a general false-positive rate on no-trigger videos. In addition,  some models also retain implementation-specific limitations, including ThinkStream's trailing-block submission behavior, JoyAI's tested complete-system boundary, and the largely unparseable spoken-style output of MiniCPM-O native duplex under the text protocol. Annotation and Judge uncertainty, timing coverage, and configuration-specific details are documented in Section~\ref{sec:benchmark} and the Appendix.

\paragraph{Future directions.}
TRACE currently prioritizes explicit temporal and execution conditions over breadth of modalities and interaction patterns. Natural extensions include no-trigger and negative-event coverage, higher visual sampling rates, audio and multi-turn interaction, and sustained-operation tests. These extensions would broaden the operating regimes covered by the benchmark while preserving the requirement that reported results remain tied to their temporal validity, execution conditions, and measurement boundaries.
\FloatBarrier
\footnotesize

\bibliographystyle{ieeenat_fullname}
\bibliography{references}
\normalsize
\appendix
\section{Related-Work Scope and Proactive Judge Calibration}

\subsection{Released Streaming-Evaluation Projects}
\label{subsec:release-landscape}

This survey compares only projects that supplied public data or annotations together with executable evaluation assets at the time of our 2026-07 release audit; projects whose artifacts were not then publicly available (e.g., EGOSTREAM, StreamingEval, VSAS-Bench) remain cited for their ideas but are not treated as runnable comparisons. Open-source availability does not imply the same task scope as TRACE. Their relationship to this report is complementary rather than uniform: StreamingBench \citep{streamingbench2024} and OVO-Bench \citep{ovobench2025} supplied the public videos and candidate tasks that TRACE re-audits; RIVER \citep{river2026} and S-EMBER \citep{sember2026} motivated reviewed evidence distances rather than source-defined difficulty labels; the proactive-response projects (ProactiveVideoQA \citep{proactivevideoqa2025}, OmniPro \citep{omnipro2026}, IPIBench \citep{ipibench2026}, ESTP-Bench \citep{estpbench2025}) establish that answer content, timing, and repetition are distinct constructs that a single hit rate cannot summarize; and the omni-modal interaction projects (OmniMMI \citep{omnimmI2025}, OmniInteract \citep{omniinteract2026}) mark the audio and multi-turn boundary beyond TRACE's current visual-only scope.

For Table~\ref{tab:benchmark_comparison}, a check denotes a construct explicitly represented in the released data/evaluator rather than merely discussed in a paper. ``Trigger rules'' requires instruction-dependent trigger semantics; ``Evidence timing'' requires per-question evidence localization; Proactive Accuracy requires an explicit task-quality score for proactive responses; Response timing requires response timing to enter the released proactive metric; and FA/Miss requires explicit response-selection error diagnostics, such as false-positive/false-negative or false-alarm/miss quantities, rather than only an aggregate score that implicitly penalizes them. Run-level columns require the released evaluation to report the named operational quantity. Under this rubric, Response timing is present for OVO-Bench, RIVER, ProactiveVideoQA, OmniInteract, and TRACE, while explicit FA/Miss-style diagnostics are present for OmniInteract and TRACE. Redundant/repeated output is not used as a comparison column because the released projects attach different semantics to repetition; TRACE reports it separately as a descriptive statistic. 

\subsection{Proactive Judge Calibration}
\label{subsec:judge-calibration}

Proactive tasks requiring semantic judgments use the same Judge configuration, prompt, and scale, following large-language-model-as-a-judge (LLM-as-judge) evaluation practice \citep{zheng2023judging, liu2023geval}. Inputs are task type, question or instruction, reference answer, and response text, without video frames. Response timing is assessed separately. Scores range from zero to one and retain partial credit. Judge request or parsing failures are recorded separately rather than presented as completed content judgments.

Initial calibration sampled 36 model responses corresponding to 31 distinct semantic items, with model identities and machine scores hidden from the human rater. Human and Judge scores were binarized at score ≥ 0.7, yielding 30/36 agreements (83.3\%; Cohen's kappa 0.675 \citep{cohen1960}): 16 true positives, 14 true negatives, zero false positives, and six false negatives, using human judgments as the reference. Thus a false negative is human-positive but Judge-negative. Six responses were drawn from each of six strata formed by Sequential Steps Recognition (SSR) and Clues Reveal Responding (CRR) tasks and machine zero/partial/full credit. This is not a simple random sample of the natural candidate distribution and is not stratified by model, so it cannot test uniformity of bias across models. Continuous-score Pearson correlation is 0.830 and Spearman correlation 0.877; the sample mean Judge-minus-human difference is −0.153, used only diagnostically. Disagreements are conservative in this sample, not evidence of equal underestimation across models or all responses.

All semantic Proactive judgments use Qwen3.5-35B-A3B with prompt version \texttt{osb-vlm-judge-v1}, temperature 0, and a 60-s request timeout through an OpenAI-compatible chat-completions endpoint. The request explicitly sets only model, temperature, and the benchmark prompt/messages; other decoding controls use the serving endpoint defaults. The exact prompt and scoring code are released with the project. Service-name and legacy metadata labels in run manifests are provenance fields rather than distinct Judge configurations. Without video input, the Judge can assess semantic agreement with reference text but cannot detect visually contradicted answers that appear textually correct. This single-rater, single-pass check provides preliminary human comparison, not complete ground truth, and does not exclude output-style bias. Binary agreement does not establish calibration of continuous partial credit. No uniform score correction is applied. Independent audits and repeated Judge tests will supply additional calibration evidence.

\subsection{Proactive Trigger Examples}
\label{subsec:trigger-examples}

Proactive annotations distinguish four instruction-dependent trigger bases. They describe when an answer becomes eligible, not task difficulty or ability levels. One video can contain windows of different types. Table~\ref{tab:trigger_examples} gives representative instructions and contrasts valid responses with false alarms that occur before the relevant response opportunity becomes valid.

\begin{center}
\begin{minipage}{\columnwidth}
\centering
\footnotesize
\setlength{\tabcolsep}{3pt}
\renewcommand{\arraystretch}{1.15}
\captionof{table}{Instruction-dependent trigger examples.}\label{tab:trigger_examples}
\begin{tabular}{@{}
  >{\raggedright\arraybackslash}p{0.30\columnwidth}
  >{\raggedright\arraybackslash}p{0.64\columnwidth}@{}}
\toprule\noalign{}
Type / annotated time & Instruction, valid response, false alarm \\
\midrule\noalign{}

Event onset \newline {\scriptsize first visible moment of the target action} &
{\scriptsize ``Alert me when someone starts opening the door.'' Valid: ``Someone is opening the door,'' after opening starts. False alarm: predicting opening before the action appears.} \\

Event completion \newline {\scriptsize completion of the target action} &
{\scriptsize ``Tell me after the door is fully closed.'' Valid: ``The door is closed,'' after closure. False alarm: saying it is closed while it is moving.} \\

Sufficient evidence \newline {\scriptsize earliest evidence supporting a unique answer} &
{\scriptsize ``State the object's color once there is enough evidence.'' Valid: ``Red,'' when color becomes identifiable. False alarm: guessing from an ambiguous outline.} \\

State interval \newline {\scriptsize from state validity until replacement} &
{\scriptsize ``Report while someone is standing.'' Valid: ``This person is standing,'' within that interval. False alarm: reporting standing before the state holds.} \\

\bottomrule
\end{tabular}
\end{minipage}
\end{center}

Point events start their windows at the trigger; state tasks end at their annotated state boundary. Onset and completion may be different moments of the same action and are selected according to instruction semantics. Outside-window responses measure temporal deviation from the instruction, not errors or false positives in a general narration task.

\section{Supplementary Evaluation Details and Results}

\subsection{QA Answer Parsing}
\label{subsec:qa-parsing}

QA Accuracy accepts a unique explicit option label, including an answer wrapper or label-prefixed option text. Unlabeled free text is not matched semantically to options. Two distinct option labels invalidate the response, even if one is mentioned only to reject it; repeating the same label does not. Ordinary articles are not option labels. Parsing uses visible answers after removing hidden thinking spans. Table~\ref{tab:qa_parsing_examples} gives representative accepted and invalid outputs.

\begin{center}
\begin{minipage}{\columnwidth}
\small
\setlength{\tabcolsep}{4pt}
\renewcommand{\arraystretch}{1.15}
\captionof{table}{Examples of recoverable QA parsing.}\label{tab:qa_parsing_examples}
\begin{tabular}{@{}
  >{\raggedright\arraybackslash}p{(\columnwidth - 24pt - 2\tabcolsep) * \real{0.50}}
  >{\raggedright\arraybackslash}p{(\columnwidth - 24pt - 2\tabcolsep) * \real{0.50}}@{}}
\toprule\noalign{}
Output & Result \\
\midrule\noalign{}

C / C. option text / The answer is C. & C \\
C. A person walks. & C; the article A is not an option label \\
A. red or B. blue & Invalid; zero credit \\
C. chosen. A. is incorrect. & Invalid; zero credit \\
Option text without a label & Invalid; zero credit \\

\bottomrule
\end{tabular}
\end{minipage}
\end{center}

Strict single-label accuracy is a format diagnostic, not the main QA metric. All main QA results use the same recoverable parsing rule.

\subsection{Supplementary Timing, Workload, and Frame Processing}
\label{subsec:timing-workload}

Table~\ref{tab:ttft} supplements the Response Latency medians in Section~\ref{sec:experiments} with time to first token (TTFT). Both measurements start at question arrival \(r_q\): TTFT ends at the first output token, whereas Response Latency ends when call completion is received. Query-time history reconstruction and input preparation after \(r_q\) are therefore included in both measurements.

\begin{center}
\begin{minipage}{\columnwidth}
\footnotesize
\setlength{\tabcolsep}{3pt}
\renewcommand{\arraystretch}{1.12}
\captionof{table}{Recorded TTFT for QA.}\label{tab:ttft}
\begin{tabular}{@{}lrrr@{}}
\toprule\noalign{}
Model/system & TTFT p50 (ms) & TTFT p95 (ms) & Field coverage \\
\midrule\noalign{}

AURA & 797.5 & 4016.4 & 99.40\% \\
MOSS-VL & 188.5 & 377.6 & 99.28\% \\
MOSS-Preview & 94.3 & 142.6 & 100.00\% \\
LiveCC & 105.1 & 130.3 & 100.00\% \\
ThinkStream & 143.6 & 1024.5 & 100.00\% \\
VideoLLM-Online & 173.3 & 859.4 & 100.00\% \\
JoyAI & 845.7 & 1775.8 & 97.94\% \\
MiniCPM-O (duplex) & 539.3 & 766.4 & 6.48\% \\

\bottomrule
\end{tabular}
\par\smallskip\begin{minipage}{\linewidth}
Table note (TTFT): Quantiles use observed values; coverage is the fraction of retained query groups with TTFT, not the fraction of all QA records with successful answers. Missing times are not imputed. These values do not establish end-to-end speed rankings.
\end{minipage}
\end{minipage}
\end{center}

Table~\ref{tab:proactive_delay_coverage} reports the exact population behind each proactive Median Response Delay. ``Answered'' counts target windows with an assigned response; ``observed onset'' counts those answered windows for which the scorer retained a response-onset time. The ratio is therefore a measurement-coverage statistic, not another task-quality metric.

\begin{center}
\begin{minipage}{\columnwidth}
\footnotesize
\setlength{\tabcolsep}{3pt}
\renewcommand{\arraystretch}{1.10}
\captionof{table}{Coverage of Proactive Median Response Delay.}\label{tab:proactive_delay_coverage}
\begin{tabular}{@{}lrrr@{}}
\toprule
Model/system & Observed onset & Answered & Coverage \\
\midrule
LiveCC & 892 & 1,266 & 70.5\% \\
MOSS-VL & 304 & 670 & 45.4\% \\
MOSS-Preview & 485 & 817 & 59.4\% \\
AURA & 357 & 665 & 53.7\% \\
ThinkStream & 70 & 401 & 17.5\% \\
VideoLLM-Online & 34 & 503 & 6.8\% \\
JoyAI & 401 & 862 & 46.5\% \\
MiniCPM-O (duplex) & 108 & 439 & 24.6\% \\
\bottomrule
\end{tabular}
\par\smallskip\begin{minipage}{\linewidth}
Median Response Delay is computed only from the observed-onset column. Low coverage does not invalidate an observed median, but it limits how broadly that median characterizes the configuration's answered windows.
\end{minipage}
\end{minipage}
\end{center}

Table~\ref{tab:proactive_workload} reports the input and generation volumes behind the Proactive comparisons in Section~\ref{subsec:proactive-results}. Image submissions include repeated history; unique frames count distinct submitted observations. The groups follow Table~\ref{tab:configurations}.

\begin{center}
\begin{minipage}{\columnwidth}
\scriptsize
\setlength{\tabcolsep}{2.4pt}
\renewcommand{\arraystretch}{1.10}
\captionof{table}{Proactive workload on 407 standard records.}\label{tab:proactive_workload}
\begin{tabular}{@{}lrrr@{}}
\toprule
Model/system & \shortstack{Submitted\\images} & \shortstack{Unique\\frames} & \shortstack{Output\\tokens} \\
\midrule
AURA & 439,400 & 34,342 & 50,957 \\
LiveCC & 23,243 & 23,240 & 78,478 \\
MOSS-Preview & 34,520* & 34,520 & 193,517 \\
MOSS-VL & 34,520* & 34,520 & 74,672 \\
ThinkStream & 32,783 & 32,780 & 373,897 \\
VideoLLM-Online & 32,783 & 32,780 & 42,162 \\
JoyAI & 313,658 & 28,371 & 797,954 \\
\bottomrule
\end{tabular}
\par\smallskip\begin{minipage}{\linewidth}\footnotesize
Table note (workload): For MOSS, the submitted-image entry uses the 34,520 unique frames recorded by the local queue because repeated submission occurrences were not retained; the identical unique-frame column makes this boundary explicit. Token totals include history descriptions, reasoning, control output, and answers. Different vocabularies and per-token computation prevent interpreting them as a common compute or monetary cost.
\end{minipage}
\end{minipage}
\end{center}

QA token-field coverage over retained calls is 99.88\% for AURA, 99.54\% for LiveCC, and 93.15\% for JoyAI; Proactive coverage is 99.55\% for LiveCC and 99.94\% for JoyAI. Other configurations with token observations have 100\% coverage. MiniCPM-O native duplex token totals are computed from recorded response text. Totals sum observed generation without extrapolation; calls lost before failure may cause additional unobserved workload.

Table~\ref{tab:proactive_record_completion} reports record-level completion separately from stream-level frame processing. These denominators are intentionally not merged: record completion asks whether an evaluation record finished, whereas Table~\ref{tab:frame_processing} summarizes retained Core frame-processing observations.

\begin{center}
\begin{minipage}{\columnwidth}
\footnotesize
\setlength{\tabcolsep}{4pt}
\renewcommand{\arraystretch}{1.12}
\captionof{table}{Proactive record-level completion for the standard response tracks.}\label{tab:proactive_record_completion}
\begin{tabular}{@{}lr@{}}
\toprule
Model/system & Record completion \\
\midrule
AURA & 99.26\% \\
LiveCC & 87.71\% \\
MOSS-Preview & 100.00\% \\
MOSS-VL & 100.00\% \\
ThinkStream & 100.00\% \\
VideoLLM-Online & 100.00\% \\
JoyAI & 95.58\% \\
\bottomrule
\end{tabular}
\par\smallskip\begin{minipage}{\linewidth}
Completion is measured over the 407 Proactive records and is distinct from target-window coverage, latency-field coverage, and stream-level frame completion. The native duplex diagnostic is excluded because its retained summaries use a different judgment/turn-taking path.
\end{minipage}
\end{minipage}
\end{center}

These measurements describe actual execution of the standard records. Some runs used longer response tolerances before their outputs were scored uniformly at five seconds. Frame grouping, queue alignment, truncation, and drops also affect observed counts. Shared scoring populations therefore do not imply identical executed input budgets.

Table~\ref{tab:proactive_redundancy} reports repeated in-window output as a descriptive behavior measure. For each target window, the first response episode is not redundant; every additional episode whose onset falls in the same strict window contributes one redundant response. This quantity is not subtracted from In-window Accuracy and is not treated as an error because repetition can be useful or undesirable depending on the interaction setting.

\begin{center}
\begin{minipage}{\columnwidth}
\footnotesize
\setlength{\tabcolsep}{4pt}
\renewcommand{\arraystretch}{1.10}
\captionof{table}{Redundant in-window response episodes on the Proactive standard set.}\label{tab:proactive_redundancy}
\begin{tabular}{@{}lrr@{}}
\toprule
Model/system & Redundant episodes & Per target window \\
\midrule
AURA & 361 & 0.284 \\
LiveCC & 2,571 & 2.024 \\
MOSS-Preview & 868 & 0.683 \\
MOSS-VL & 2,357 & 1.856 \\
ThinkStream & 48 & 0.038 \\
VideoLLM-Online & 68 & 0.054 \\
JoyAI & 948 & 0.746 \\
\bottomrule
\end{tabular}
\end{minipage}
\end{center}

Table~\ref{tab:frame_processing} reports stream-level completion and dropped-frame observations used in the reliability discussion. These quantities use measurement sources distinct from record-level completion.

\begin{center}
\begin{minipage}{\columnwidth}
\footnotesize
\setlength{\tabcolsep}{3pt}
\renewcommand{\arraystretch}{1.12}
\captionof{table}{Proactive frame processing.}\label{tab:frame_processing}
\begin{tabular}{@{}lrr@{}}
\toprule\noalign{}
Model/system & Stream completion & Dropped-frame rate \\
\midrule\noalign{}

AURA & 100.00\% & 0.00\% \\
LiveCC & 70.59\% & 29.11\% \\
MOSS-Preview & 99.99\% & 0.01\% \\
MOSS-VL & 99.99\% & 0.01\% \\
ThinkStream & 99.99\% & 0.01\% \\
VideoLLM-Online & 100.00\% & 0.01\% \\
JoyAI & 86.55\% & 13.46\% \\

\bottomrule
\end{tabular}
\par\smallskip\begin{minipage}{\linewidth}
Table note (frames): Stream completion uses retained Core observations with submission-completion timestamps; dropped-frame rate uses observations without actual submission records. Different reporting sources mean these rates need not sum to 100\%. Queue submission or history insertion does not establish completed internal-state updates or on-time processing; per-frame latency and on-time rates are therefore not compared here.
\end{minipage}
\end{minipage}
\end{center}

\begin{figure*}[!b]
\centering
\includegraphics[width=0.75\textwidth]{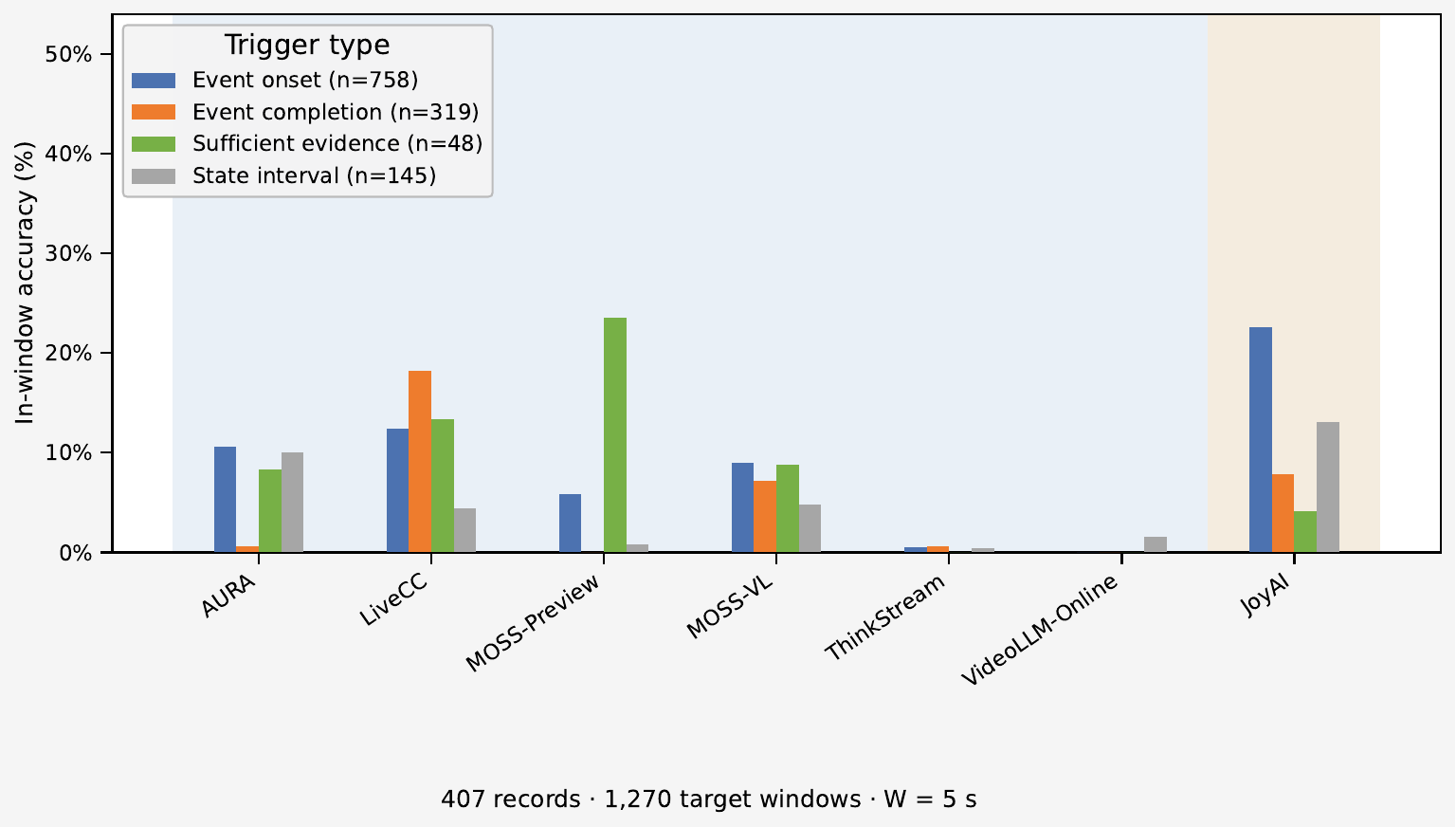}
\caption{In-window Accuracy by Proactive trigger type for seven standard response tracks. Bars are current-v4 point estimates; legend entries give target-window counts. Shading separates the standard model-side and complete-system (JoyAI) tracks. Trigger subsets differ in size and content and are descriptive rather than controlled difficulty groups.}
\label{fig:trigger_breakdown}
\end{figure*}

\subsection{Trigger-Type Results and Statistical Uncertainty}
\label{subsec:trigger-results}

Scores retain partial credit. Subsets may share videos, so video counts are not additive. Figure~\ref{fig:trigger_breakdown} visualizes trigger-type point estimates for the seven standard response tracks from the same current-v4 standard-set window analysis used for the aggregate results. The native duplex diagnostic is omitted because a matching trigger-level breakdown was not retained in the audited export bundle. Table~\ref{tab:trigger_subsets} gives exact values for AURA and the two MOSS configurations discussed in Section~\ref{subsec:proactive-results}; the AURA trigger rows aggregate to its 7.92\% overall In-window Accuracy.

\begin{center}
\begin{minipage}{\columnwidth}
\scriptsize
\setlength{\tabcolsep}{1.7pt}
\renewcommand{\arraystretch}{1.12}
\captionof{table}{Exact audited In-window Accuracy by Proactive trigger type for AURA and the two MOSS configurations.}\label{tab:trigger_subsets}
\begin{tabular}{@{}lrrrrr@{}}
\toprule\noalign{}
Trigger type & Win. & Vid. & AURA & \shortstack{MOSS-\\Preview} & MOSS-VL \\
\midrule\noalign{}

Event onset & 758 & 50 & 10.55\% & 5.80\% & 8.97\% \\
Event completion & 319 & 80 & 0.63\% & 0.00\% & 7.21\% \\
Sufficient evidence & 48 & 10 & 8.33\% & 23.54\% & 8.75\% \\
State interval & 145 & 40 & 10.07\% & 0.83\% & 4.83\% \\

\bottomrule
\end{tabular}
\end{minipage}
\end{center}

The report's video-clustered 95\% confidence intervals use 2,000 percentile-bootstrap resamples at the video level, keeping tasks from the same video together. Pairwise differences use the same sampled video clusters for both configurations on their shared-video population; deterministic seeds are derived from the compared cohort and model names. This avoids treating related records or windows as independent. Intervals describe uncertainty in point estimates; overlapping intervals do not establish a ranking.

\subsection{Response-Segment Assignment}
\label{subsec:segment-assignment}

A complete response segment is assigned by its first-token time and judged by its eventual content. Thus a segment whose onset lies outside a strict window remains outside that window even if the target content appears later during generation. Missing first-token timestamps fall back to Core receipt time and are marked incomplete for latency coverage. Additional responses after an event has received an answer are redundant even when the first answer was wrong.

Execution ends at the last strict-window boundary. If the video ends earlier, the remaining wait supplies no new frames. The auxiliary post-trigger eventual score allows point-event content credit after the strict window but before the next trigger; the last point event retains its strict boundary, and state tasks retain their annotated end. Windows without recorded latency are excluded from the Median Response Delay; failed Judge requests are distinguished from completed content judgments.

MiniCPM-O native duplex is a diagnostic configuration, receiving the instruction once and relying on listen/speak decisions; the turn-taking behavior of such duplex models is the focus of dedicated spoken-dialogue benchmarks \citep{fullduplexbench2025}. Its standard-set In-window Accuracy is 0.51\% over all 1,270 target windows (video-clustered 95\% CI 0.18--0.93\%), with a 1.24\,s median delay computed from 108 observed onsets among 439 answered windows. Its QA outputs are spoken-style fragments, and 779 of 833 answers cannot be uniquely parsed (Appendix~B.1). The diagnostic is therefore reported under its native interaction boundary rather than treated as a text-interface baseline.

\end{document}